\documentclass[letterpaper]{article} 
\usepackage{aaai24}  
\usepackage{times}  
\usepackage{helvet}  
\usepackage{amsmath}
\usepackage{tabularx}
\usepackage{booktabs}
\usepackage{subcaption}
\usepackage{courier}  
\usepackage[hyphens]{url}  
\usepackage{graphicx} 
\usepackage{natbib}  
\usepackage{caption} 
\usepackage{algorithm}
\usepackage{algorithmic}

\usepackage{newfloat}
\usepackage{listings}
\DeclareCaptionStyle{ruled}{labelfont=normalfont,labelsep=colon,strut=off} 
\floatstyle{ruled}
\newfloat{listing}{tb}{lst}{}
\floatname{listing}{Listing}
\title{Topic-VQ-VAE: Leveraging Latent Codebooks for \\Flexible Topic-Guided Document Generation}
\author{
    YoungJoon Yoo\textsuperscript{\rm 1}~
    Jongwon Choi\textsuperscript{\rm 2}
}
\affiliations{
    \textsuperscript{\rm 1} ImageVision, NAVER, Cloud.\\
    \textsuperscript{\rm 2}  Department of Advanced Imaging (GSAIM) and Graduate School of AI, Chung-Ang University.\\
    \textsuperscript{\rm 1}youngjoon.yoo@navercorp.com, 
    \textsuperscript{\rm 2}choijw@cau.ac.kr\\
}

\usepackage{bibentry}

\begin{document}

\maketitle

\begin{abstract}
This paper introduces a novel approach for topic modeling utilizing latent codebooks from Vector-Quantized Variational Auto-Encoder~(VQ-VAE), discretely encapsulating the rich information of the pre-trained embeddings such as the pre-trained language model.
From the novel interpretation of the latent codebooks and embeddings as conceptual bag-of-words, we propose a new generative topic model called Topic-VQ-VAE~(TVQ-VAE) which inversely generates the original documents related to the respective latent codebook.
The TVQ-VAE can visualize the topics with various generative distributions including the traditional BoW distribution and the autoregressive image generation.
Our experimental results on document analysis and image generation demonstrate that TVQ-VAE effectively captures the topic context which reveals the underlying structures of the dataset and supports flexible forms of document generation.
Official implementation of the proposed TVQ-VAE is available at \url{https://github.com/clovaai/TVQ-VAE}.
\end{abstract}

\section{Introduction}

Topic modeling, the process of extracting thematic structures, called \textbf{topic}, represented by coherent word sets and subsequently clustering and generating documents based on these topics, constitutes a foundational challenge in the manipulation of natural language data  
The initiative Latent Dirichlet Allocation (LDA) \citep{blei2003latent} and subsequent studies~\cite{teh2004sharing,paisley2014nested} configure the inference process as a Bayesian framework by defining the probabilistic generation of the word, interpreted as bag-of-words (BoW), by the input word and document distributions. 
The Bayesian frameworks utilize the co-occurrence of the words in each document and have become a standard for topic models.

Despite the success, topic modeling has also faced demands for the evolution to reflect advances of recent deep generative studies. One main issue is utilizing information from large-scale datasets encapsulated in pre-trained embeddings~\cite{pennington2014glove,devlin2018bert,radford2021learning}. Many follow-up studies have approached the problem in generative~\cite{dieng2020topic} or non-generative~\cite{duan2021sawtooth,xu2022hyperminer,grootendorst2022bertopic} directions.
Moreover, with the advancements in generation methods, such as autoregressive and diffusion-based generation, there is a growing need for the topic-based generation to evolve beyond the traditional BoW form and become more flexible.

To address the issue, we propose a novel topic-driven generative model using Vector-Quantized~(VQ) embeddings from \cite{van2017neural}, an essential building block for the recent vision-text generative model such as \cite{ramesh2021zero}.
In contrast to previous approaches in topic modeling~\cite{gupta2021vector,gupta2023neural} that treat VQ embeddings as topics, in our method, each VQ embedding represents the embeddings of conceptually defined words.
Through the distinct perspective, we achieve the enhanced flexibility that a corresponding codebook serves as its BoW representation.
We further demonstrate that the codebook consisting of VQ embedding itself is an implicit topic learner and can be tuned to achieve exact topic context, with a supporting flexible format of sample generation.

Based on the interpretation, we present a novel generative topic model, Topic-VQ Variational Autoencoder (TVQ-VAE), which applies a VQ-VAE framework~\cite{van2017neural} incorporating topic extraction to the BoW representation of the VQ-embedding.
The TVQ-VAE facilitates the generation of the BoW-style documents and also enables document generation in a general configuration, simultaneously.
We demonstrate the efficacy of our proposed methodology in two distinct domains: (1) document clustering coupled with set-of-words style topic extraction, which poses a fundamental and well-established challenge in the field of topic modeling. For the pre-trained information, we utilize codebooks derived from inputs embedded with a Pre-trained Language Model (PLM)~\cite{reimers2019sentence}. Additionally, (2) we delve into the autoregressive image generation, leveraging the VQ-VAE framework with latent codebook sequence generation as delineated in  \cite{van2016pixel,esser2021taming}.

The contributions of the paper are summarized as follows: \begin{itemize}
    \item We propose a new generative topic modeling framework called \textbf{TVQ-VAE} utilizing codebooks of VQ embeddings and providing a flexible form of sampling. 
    Our proposed model interprets the codebooks as a \textbf{conceptual} word and extracts the topic information from them.
    \item Our proposed model \textbf{TVQ-VAE} provides a general form of probabilistic methodology for topic-guided sampling. We demonstrate the application of samplings, from a typical histogram of the word style sample used in the topic model to an autoregressive image sampler.
    \item 
    From the extensive analysis of two different data domains: (1) document clustering typically tackled by the previous topic models and (2) autoregressive image generation with topic extraction. The results support the proposed strength of the \textbf{TVQ-VAE}.
\end{itemize}

\section{Preliminary}
\subsection{Key Components of Topic Model} 
\label{prelim:topic}
We summarize the essence of the topic model where the generative or non-generative approaches commonly share as (1) semantic topic mining for entire documents and (2) document clustering given the discovered topics. 
Given $K$ number of topics $\beta_k\in\boldsymbol{\beta}, k=1,...,K$, the topic model basically assigns the document to one of $K$ topics, which is a clustering process given the topics.
This assigning can be deterministic or generatively by defining the topic distribution of each document, as:
\begin{eqnarray}
\begin{aligned}
\label{eq:d_t}
z_{dn} \sim p(z| \theta_d),
\end{aligned}
\end{eqnarray}
where the distribution $p(z| \theta_d)$ draws the indexing variable $z_{dn}$ that denotes the topic index $\beta_{z_{dn}}$ that semantically includes the word $w_{dn}$ in $d$'th document. In a generative setting, the random variable $\boldsymbol{\theta}$ is typically defined as $K$ dimensional Categorical~\cite{blei2003latent} distribution with Dirichlet prior $\alpha$ or Product of Expert (PoE)~\cite{srivastava2017autoencoding}.
The topic $\beta_k$ is defined as a set of semantically coherent words $w_{kn}\in\beta_k, 1,..., N_w$ or by a word distribution in a generative manner, as:
\begin{eqnarray}
\begin{aligned}
\label{eq:w_t}
w_k \sim p(w | \beta_k).
\end{aligned}
\end{eqnarray}
Similarly, the $p(w | \beta_k)$ can be defined as categorical~\cite{blei2003latent} like distributions.
Classical probabilistic generative topic models~\cite{blei2003latent,srivastava2017autoencoding,miao2016neural,zhang2018whai,nan2019topic} interpret each document $d$ as BoW $\mathbf{w}_d=\{w_{d1},...,w_{dn}\}$ and analysis the joint distribution $p(\boldsymbol{\theta}, \boldsymbol{\beta}|\mathbf{w}_d)$ from equations~(\ref{eq:d_t}-\ref{eq:w_t}), by approximated Bayesian inference methods~\cite{casella1992explaining,wainwright2008graphical,kingma2013auto}.
We note that their probabilistic framework reflects word co-occurrence tendency for each document.

When embedding is applied to the topic modeling frameworks~\cite{dieng2020topic,duan2021sawtooth,xu2022hyperminer,meng2022topic},
some branches of embedded topic models preserve the word generation ability, and hence the word embedding is also included in their probabilistic framework, such as ETM~\cite{dieng2020topic}. 
The non-generative embedded topic models including recent PLM-based topic models~\cite{sia2020tired,grootendorst2022bertopic,meng2022topic} extract topic embedding directly from distance-based clustering method, bypassing the complicated Bayesian inference approximation, with utilizing in post-processing steps.

\subsection{Vector Quantized Embedding} 
Different from the typical autoencoders mapping an input $x$ to a continuous latent embedding space $\mathcal{E}$, Vector-Quantized Variational Auto-Encoder (VQ-VAE)~ \cite{van2017neural} configures the embedding space to be discrete by the VQ embeddings $\boldsymbol\varrho=\{\rho_n\in \mathcal{R}^{D_{\rho}}, n=1,..., N_{\rho}\}$.
Given the encoder function of the VQ-VAE as $f = Enc(x; W_E)$, the vector quantizer $(c_x,\rho_x)=Q(f)$ calculates the embedding $\rho_x\in\boldsymbol\varrho$, which is the closest embedding to $f$ among the set of VQ embedding $\boldsymbol\varrho$, and its one-hot encoded codebook $c_x\in\mathcal{R}^{N_{\rho}}$. The embedding $\rho_x$ and $c_x$ is defined as:  
\begin{eqnarray}
\begin{aligned}
\label{eq:vq_onehot}
&\rho_x = c_x\cdot\hat{\boldsymbol{\rho}},~
&\hat{\boldsymbol{\rho}} = [\rho_1,...,\rho_{N_{\rho}}]\in\mathcal{R}^{N_{\rho}\times D_{\rho}},
\end{aligned}
\end{eqnarray}
where $N_\rho$ denotes the size of the discrete latent space, which is smaller than the original vocabulary size $N_w$. 
$D_\rho$ is the dimensionality of each latent embedding vector.
Here, we denote the resultant sets of embedding $\boldsymbol\rho$ and codebook $\boldsymbol{c}$ are defined as $\boldsymbol\rho=\{\rho_x\}$ and $\boldsymbol{c}=\{c_x\}$. 
When given an image $x\in\mathcal{R}^{H\times W\times 3}$ as a VQ-VAE input, we collect the sequence of quantized vector $\boldsymbol{\rho}$ and $\boldsymbol{c}$
 as:
 \begin{eqnarray}
\begin{aligned}
\label{eq:vq_set}
\boldsymbol{\rho}&=\{\rho_{ij}\in\boldsymbol\varrho| i=1,...,h, j=1,...,w\},\\
\boldsymbol{c} &= \{c_{ij}\in\mathcal{R}^{N_\rho}| i=1,...,h, j=1,...,w\}, 
\end{aligned}
\end{eqnarray}
where the embedding $\rho_{ij}$ and the codebook $c_{ij}$ maps the closest encoding of the spatial feature $f_{ij}$ of the latent variable $\boldsymbol{f}=\{f_{ij}|i=1,...,h, j=1,...,w\}, \boldsymbol{f} = Enc(x;W_E)\in\mathcal{R}^{h\times w\times d}$.
The decoder function $\tilde{x} = Dec(\boldsymbol{c}, \boldsymbol\rho; W_D)$ then reconstruct the original image $x$ using the VQ embedding $\boldsymbol\rho$ and its codebook $\boldsymbol{c}$. In this case, the vector quantizer $Q(\cdot)$ calculates the sequence of codebook $\boldsymbol{c}$ and the corresponding embeddings $\boldsymbol{\rho}$, as $(\boldsymbol{c}. \boldsymbol\rho)=Q(\boldsymbol{f})$.

\section{Methodology}
We present a new topic-driven generative model, TVQ-VAE, by first introducing a new interpretation to the VQ-VAE output: codebooks $\boldsymbol{c}$ and their embedding $\boldsymbol\rho$.

\begin{figure*}[t]
\begin{center}
    \begin{subfigure}[t]{0.23\linewidth}
        \includegraphics[width=0.95\linewidth]{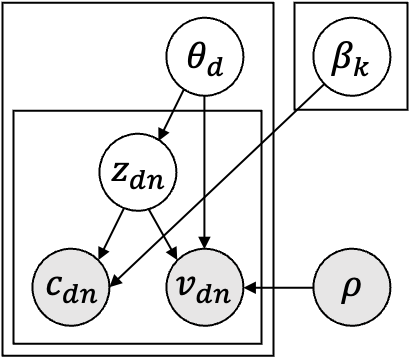}
        \centering
        \caption{BoW form.}
        \label{fig:graphical_model_bow}
    \end{subfigure}
    \begin{subfigure}[t]{0.23\linewidth}
        \includegraphics[width=0.95\linewidth]{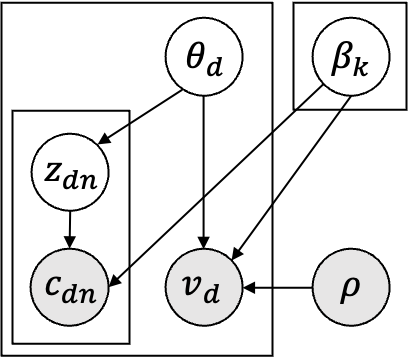}
        \centering
        \caption{General form.}
        \label{fig:graphical_model_general}
    \end{subfigure}    
    \begin{subfigure}[t]{0.49\linewidth}
        \includegraphics[width=0.99\linewidth]{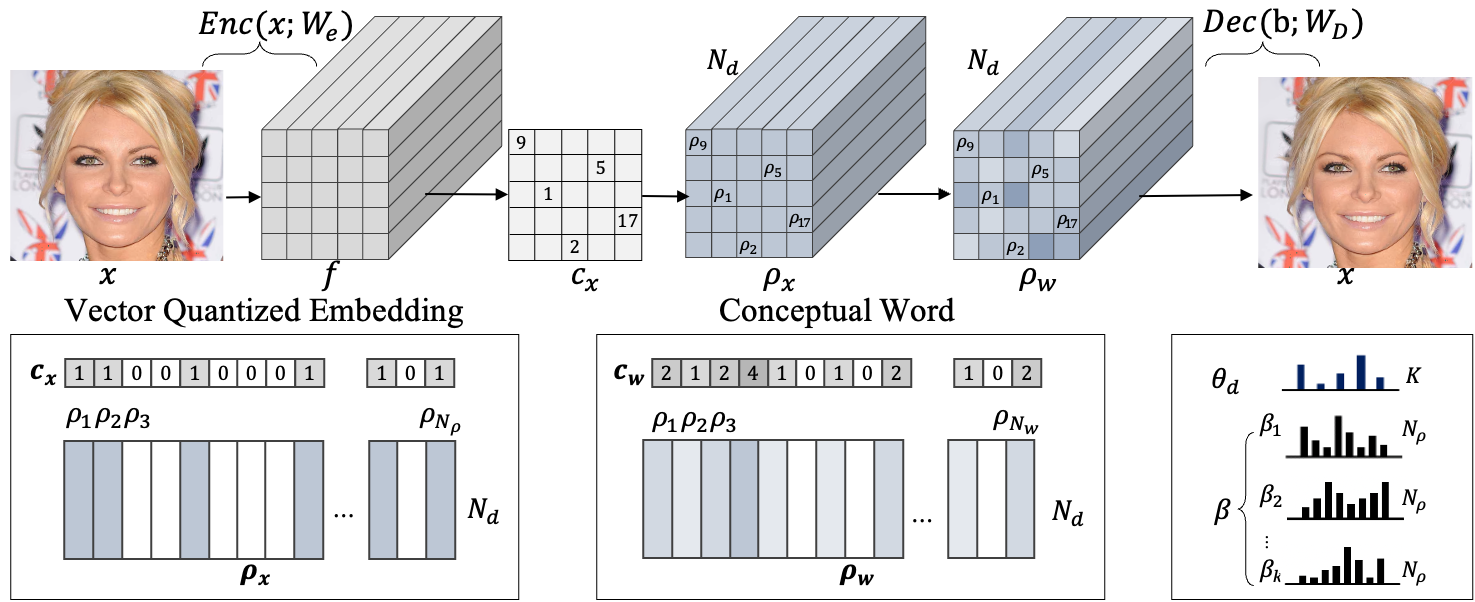}
        \centering
        \caption{Visualized diagram of TVQ-VAE.}
        \label{fig:graphical_model_diagram}
    \end{subfigure}
\end{center}
\caption{Graphical representation of the TVQ-VAE. Diagrams (a) and (b) illustrate the TVQ-VAE's graphical representation in both BoW and General forms, while diagram (c) presents an example of vector quantized embedding, conceptual word, and output. Notably, the encoder network is fixed in our method.}
\label{fig:graphical_model}
\end{figure*}

\subsection{Vector Quantized Embedding as Conceptual Word}
\label{sec:BoWE}
Here, we first propose a new perspective for interpreting a set $\boldsymbol{B}$ including the VQ embedding $\rho$ and its codebook $c$: 
\begin{eqnarray}
\begin{aligned}
\label{eq:vq_onehot_2}
\boldsymbol{B} =\{b_i = (c_i, \rho_i) | i=1,...N_{\rho
}\},
\end{aligned}
\end{eqnarray}
as \textbf{conceptual} word. The conceptual word $b_i$ each consists of VQ embedding $\rho_i$ and its codebook $c_i$. We note that the number of the virtual word $b_i$ is equivalent to total number $N_{\rho}$ of VQ embeddings. 

One step further, since the typical selection of the number $N_{\rho}$ is much smaller than the original vocabulary, we modify the set $B$ so that multiple embeddings express the input, where the codebook $c$ in Equation~(\ref{eq:vq_onehot}) becomes a multi-hot vector. This relaxation lets the codebooks deal with a much larger size of words. 
Specifically, given word $w$ and its embedding $z_w = Enc(w)$ from the VQ-VAE encoder, we support the expansion from one-hot to multi-hot embedding by using $K$-nearest embeddings $\rho_1,...,\rho_k$ from $B$ to represent quantized embedding $\rho_w $ for $z_w$ as:
\begin{eqnarray}
\begin{aligned}
\label{eq:vq_multihot}
c_w = \sum_k {c}_k,\\
\rho_w = c_w\cdot\hat{\boldsymbol{\rho}},
\end{aligned}
\end{eqnarray}
where the matrix $\hat{\boldsymbol{\rho}}$ denotes the encoding matrix in Equation~(\ref{eq:vq_onehot}). Using the expanded codebook $c_w$ and its embedding $\rho_w$ from equation~(\ref{eq:vq_multihot}), we define a expanded Bag-of-Word $\boldsymbol{B}_w$, the final form of the \textbf{conceptual} word, as follows: 
\begin{eqnarray}
\begin{aligned}
\label{eq:BoWE}
\boldsymbol{B}_w=\{b_w = (c_{w}, \rho_{w}) | w=1,...,N_{w} \}.
\end{aligned}
\end{eqnarray}
We note that the multi-hot embedding $c_{w}\in\mathcal{R}^{N_\rho}$ is defined as $N_\rho$ dimensional vector which is $N_{w}>>N_{\rho}$.
Theoretically, the cardinality of $\boldsymbol{B}_w$ increases to combinatorial order $\binom{N_{\rho}}{K}$, where the number $K$ called expansion value, denotes the number of assigned embeddings for each input.

\subsection{Generative Formulation for TVQ-VAE}
This section proposes a generative topic model called TVQ-VAE analyzing the \textbf{conceptual} words $\boldsymbol{B}_w$ in Equation~(\ref{eq:BoWE}). As illustrated in the graphical model in Figure~\ref{fig:graphical_model}, the TVQ-VAE model follows typical topic modeling structures formed by independent $d=1,...,D$ documents, and each document $d$ has independent $N_w$ words $c_{w}\equiv c_{dn}\in\mathcal{R}^{N_{w}}$, $n=1,..., N_w$. An output sample $v_d$ is matched to a document $d$.
TVQ-VAE provides various output forms for $v_d$. For the typical set-of-word style output, $v_d$ is defined as a set of word $v_d = \{v_{d1},...,v_{dN_w}\}$ (Figure~\ref{fig:graphical_model_bow}), where the word $v_{dn}\in\mathcal{R}^{N_w}$ denotes the one-hot encoding of the original word $w_{dn}$ corresponding to $c_{dn}\in\mathcal{R}^{N_\rho}$. Also, we can define $v_d$ as an image corresponding to the document $d$ (Figure~\ref{fig:graphical_model_general}).

The joint distribution of the overall random variable $\{\boldsymbol{\theta}, \boldsymbol{z}, \boldsymbol{v}, \boldsymbol{c}, \boldsymbol{\beta}, \boldsymbol{\rho}\}$ is formulated as:
\begin{eqnarray}
\begin{aligned}
\label{eq:tvq_vae_joint}
&p(\boldsymbol{\theta}, \boldsymbol{z}, \boldsymbol{v}, \boldsymbol{c}, \boldsymbol{\beta}, \boldsymbol{\rho}) \\
&= 
p(\boldsymbol{\theta},\boldsymbol{\beta},\boldsymbol{\rho})\prod^{D}_{d=1}p(v_d|\theta_d, \boldsymbol{\beta}, \boldsymbol{\rho})\prod^{N_w}_{n=1}p(c_{dn}|\beta{z_{dn}})p(z_{dn}|\theta_d),
\end{aligned}
\end{eqnarray}
where the distribution $p(\boldsymbol{\theta},\boldsymbol{\beta},\boldsymbol{\rho})$ denotes the prior distribution for each independent random variable. 
The configuration in Equation~(\ref{eq:tvq_vae_joint}) is a typical formulation for the generative topic model from \cite{blei2003latent} or \cite{dieng2020topic}, each defines $p(c|\beta_{z_{dn}})$ and $p(z_{dn}|\theta_d)$ to be \textbf{categorical} and \textbf{softmax} distribution. The main factor that discriminates the previous topic models to TVQ-VAE here is the generation of the output $v_d$ from $p(v_d|\theta_d, \boldsymbol{\beta}, \boldsymbol\rho)$. 

As mentioned above, TVQ-VAE supports various forms of generation for output $v_d$.
First, for the typical set-of-word style output $v_d = \{v_{d1},...,v_{dN_w}\}$, as in Figure~\ref{fig:graphical_model_bow}, the generation $p(v_d|\theta_d, \boldsymbol\beta, \boldsymbol\rho)$ is defined as:
\begin{eqnarray}
\begin{aligned}
\label{eq:tvq_gen_sow}
p(v_d|\theta_d, \boldsymbol\beta, \boldsymbol\rho)=\prod^{N_w}_{n=1}\sum^{K}_{z_{dn}=1}p(v_{dn}|\alpha(\beta_{z_{dn}}\cdot\hat{\boldsymbol{\rho}}))p(z_{dn}|\theta_d),
\end{aligned}
\end{eqnarray}
where a trainable fully connected layer  $\alpha\in\mathcal{R}^{N_w\times N_{\rho}}$ connects the topic embedding $\beta_{z_{dn}}\cdot\hat{\boldsymbol\rho}\in\mathcal{R}^{N_{\rho}}$ to the original word dimension. 
Here, we define $p(v|\cdot)$ and $p(z_{dn}|\cdot)$ as \textbf{softmax} distribution, which is a PoE implementation of the topic model in \cite{srivastava2017autoencoding}.
We note that it is possible to priorly marginalize out the indexing variable $z_{dn}$ in equation~(\ref{eq:tvq_gen_sow}) by computing all the possible cases of sample drawn from $p(z_{dn}|\theta_d)$.

\begin{algorithm}[t]
\caption{Pseudo-code of TVQ-VAE generation}
\label{alg:gen}
\begin{algorithmic}[1] 
\REQUIRE Given an topics $\boldsymbol\beta=\{\beta_1,...,\beta_K\}$,
\STATE Sample or define $\theta_d$.
\IF{document analysis}
    \STATE Sample $z_{dn}\sim p(z|\theta_d)$: $p(z|\cdot)$ be the softmax. 
    \STATE $v_{dn} \sim p(v|\alpha(\beta_{z_{dn}}\cdot\hat{\boldsymbol\rho}))$: $p(v|\cdot)$ be the softmax.     
    \STATE \textbf{Repeat} $n=1,...,N_w$
\ELSIF{Image generation}
    \STATE $\textbf{c}'\sim$ AR($\boldsymbol\theta\cdot\hat{\boldsymbol\beta}\cdot\hat{\boldsymbol\rho}$).
    \STATE $v = Dec(\textbf{c}',\boldsymbol\rho)$, $Dec(\cdot)$ be VQ-VAE decoder.
\ENDIF
\end{algorithmic}
\end{algorithm}
\begin{algorithm}[t]
\caption{Pseudo-code of TVQ-VAE training}
\label{alg:train}
\begin{algorithmic}[1] 
\REQUIRE The batch of the input $x_d$ and the output $v_d$. 
\IF{document analysis}
    \STATE $x_d$ is the PLM vector from each Sentence.
    \STATE $v_d$ be the histogram of the original word.
\ELSIF{Image generation}
    \STATE $x_d\in\mathcal{R}^{H\times W\times 3}$ is an image.
\ENDIF
\STATE Initialize $\boldsymbol\beta$, $\gamma_p$.
\STATE $(\boldsymbol\rho, \boldsymbol{c}) = Q(Enc(x; W_{E})).$
(In equation~(\ref{eq:vq_onehot}-\ref{eq:vq_set}) and (\ref{eq:vq_multihot})).
\STATE Calculate $\theta$ from $q(\theta|\gamma)$ (In equation~(\ref{eq:elbo})).\\
\STATE \hspace{0.3cm}$(\gamma_m, \log(\gamma_\sigma)) = NN(\textbf{c}; W_\gamma)$.
\STATE \hspace{0.3cm}$\theta_d = Reparam(\gamma_m, \log(\gamma_\sigma))$.
\IF{document analysis}
    \STATE $\boldsymbol\beta = \alpha(\theta_d\cdot\hat{\boldsymbol\beta}\cdot\hat{\boldsymbol\rho})$.
\ELSIF{Image generation}
    \STATE $\textbf{c}'$ = AR($\theta_d\cdot\hat{\boldsymbol\beta}\cdot\hat{\boldsymbol\rho}; W_{ar}$).
\ENDIF
\STATE $l_{KL} = D_{KL}(\log(\gamma_\sigma), \gamma_m, \gamma_p)$.
\STATE $l_c = \textbf{c}*\log(softmax(\theta_d\cdot\hat{\boldsymbol\beta}))$.
\IF{document analysis}
    \STATE $l_v = v_d*\log(\beta)$.
\ELSIF{Image generation}
    \STATE $l_v = CE(\textbf{c}, \textbf{c}')$.
\ENDIF 
\STATE $l = l_{KL} +l_c + l_v$.
\end{algorithmic}
\end{algorithm}

For a more general case (Figure~\ref{fig:graphical_model_general}), we assume the output $v_{d}$ is generated by a sequence of codebook $\textbf{c}_d=\{c_{dn} |n=1,..N_w\}$ and VQ-VAE decoder $v_d=Dec(\boldsymbol{c}_d, \boldsymbol\rho; W_D)$. To generate $\textbf{c}_d$, we use AR prior $p_{ar}(\cdot)$ including PixelCNN and Transformer~\cite{esser2021taming}, as:
\begin{eqnarray}
\begin{aligned}
\label{eq:tvq_gen_ar}
p(v_d&=Dec(\textbf{c}_d,\boldsymbol\rho_d)|\theta_d, \boldsymbol\beta, \boldsymbol\rho) = P(\textbf{c}_d|\theta_d\cdot\hat{\boldsymbol\beta}\cdot\hat{\boldsymbol\rho}) \\
&=\prod^{N}_{n=1}p_{ar}(c_{dn}|c_{dn-1},...,c_{d1}, \theta_d\cdot\hat{\boldsymbol\beta}\cdot\hat{\boldsymbol\rho}),
\end{aligned}
\end{eqnarray}
where the matrix $\hat{\boldsymbol\beta}$ denotes $\hat{\boldsymbol\beta}=[\beta_1,...,\beta_K]$.
We note that $Dec(\cdot)$ is a deterministic function, and the AR prior coupled with VQ-VAE decoding provides Negative Log Likelihood (NLL)-based convergence to the general data distributions. A detailed explanation of the generation algorithm is given in Algorithm~(\ref{alg:gen}).

\subsection{Training TVQ-VAE}
For the model inference, we leverage autoencoding Variational Bayes (VB)~\cite{kingma2013auto} inference to the distribution in Equation~(\ref{eq:tvq_vae_joint}) in a manner akin to \cite{srivastava2017autoencoding,dieng2020topic}.
In these methods, VB inference defines the variational distribution $q(\boldsymbol\theta, \boldsymbol{z}|\gamma, \phi)$ that can break the connection between $\boldsymbol\theta$ and $\boldsymbol{z}$, as $q(\boldsymbol\theta, \boldsymbol{z}|\gamma, \phi)=q(\boldsymbol\theta|\gamma)q(\boldsymbol{z}|\phi)$, of the posterior distribution $p(\boldsymbol\theta, \boldsymbol{z}|\mathbf{c},\mathbf{v},\beta,\rho)$.
By the VB, the ELBO here is defined as:
\begin{eqnarray}
\begin{aligned}
\label{eq:elbo}
L(\gamma)=&-D_{KL}[q(\boldsymbol\theta|\gamma)||p(\boldsymbol\theta)]\\
&+E_{q(\boldsymbol\theta|\gamma)}[\log p(\mathbf{c},\mathbf{v}|\boldsymbol{z},\boldsymbol\theta,\boldsymbol\rho, \boldsymbol\beta)],
\end{aligned}
\end{eqnarray}
where we pre-marginalize out the $\boldsymbol{z}$, similar to equation~(\ref{eq:tvq_gen_sow}).
In the equation, the first term measures the Kullbeck-Leibler (KL) distance between the variational posterior over the real posterior distribution, called \textbf{KL} term, and the second term denotes the \textbf{reconstruction} term.
Followed by \cite{dieng2020topic}, we define the variational parameter $\gamma=NN(\boldsymbol{c}; W_\gamma)$ as a neural network (NN) function of the input set-of-word $\boldsymbol{c}$, and $\boldsymbol\theta$ is drawn by a reparameterization technique given the variable $\gamma$.

Different from the previous methods~\cite{srivastava2017autoencoding,dieng2020topic}, we also consider the reconstruction of the output samples $\textbf{v}$, as:
\begin{eqnarray}
\begin{aligned}
\label{eq:reconstruction}
&E_{q_{\gamma}(\theta)}[\log p(\mathbf{c},\mathbf{v}|\boldsymbol{z},\boldsymbol\theta,\boldsymbol\rho, \boldsymbol\beta)] = \\
&E_{q_{\gamma}(\theta)}[\log p(\mathbf{c}|\boldsymbol{z},\boldsymbol\theta,\boldsymbol\rho, \boldsymbol\beta)] 
+ E_{q_{\gamma}(\theta)}[\log p(\mathbf{v}|\boldsymbol{z},\boldsymbol\theta,\boldsymbol\rho, \boldsymbol\beta)].
\end{aligned}
\end{eqnarray}
Here, $\mathbf{c}$ and $\mathbf{v}$ are conditionally independent given $\theta$, as in Figure~\ref{fig:graphical_model_general}.
Therefore, the TVQ-VAE model has three loss terms corresponding to KL and the reconstruction terms:
\begin{eqnarray}
\begin{aligned}
\label{eq:loss}
l_{tot} = l_{KL}(\theta) + l_{rec}(\mathbf{c}) + l_{rec}(\mathbf{v}).
\end{aligned}
\end{eqnarray}
\paragraph{Training Implementation. }Since the KL divergence calculation in equation~(\ref{eq:loss}), which is $l_{KL}(\theta)$, and the first term in equation~(\ref{eq:reconstruction}), which is $l_{rec}(\mathbf{c})$, is equivalent to the VB calculation of the classical topic model, we employ the Prod-LDA setting in \cite{srivastava2017autoencoding} to those terms. For the last reconstruction term $l_{rec}(\mathbf{v})$, we can use the generative distributions defined in Equation~(\ref{eq:tvq_gen_sow}) for a set-of-word style document $v_d$ or autoregressive generation given PixelCNN prior as in Equation~(\ref{eq:tvq_gen_ar}).
We note that for the first case, the reconstruction loss term has equivalent to those of the reconstruction term for $\mathbf{c}$, and for the second case, the loss term is equivalent to the AR loss minimizing NLL for both PixelCNN and Transformer.
A detailed training process is given in Algorithm~(\ref{alg:train}).

The overall trainable parameters for the topic modeling in the process are $W_\gamma$ for the variational distribution $\gamma$, the topic variable $\boldsymbol\beta$. For the sample generation, the feedforward network $\alpha(\cdot)$ and AR parameter $W_{ar}$ are also trained for document analysis and image generation cases. It is possible to train VQ-VAE encoder $W_E$ as well, but we fix the VQ-VAE parameters considering that many studies utilize pre-trained VQ-VAE without scratch training.

\section{Related Works}
Since the initiative generative topic modeling from \cite{blei2003latent}, many subsequent probabilistic methods~\cite{teh2004sharing,paisley2014nested} have been proposed.
After the proposal of autoencoding variational Bayes, a.k.a., variational autoencoder (VAE), from ~\cite{kingma2013auto}, neural-network-based topic models (NTMs)~\cite{miao2016neural,srivastava2017autoencoding,zhang2018whai,nan2019topic} have been proposed. 
To reflect the discrete nature of the topic, \cite{gupta2021vector,gupta2023neural} introduces discrete inference of the topics by VQ-VAE \cite{van2017neural}.
Unlike the above methods that treat each Vector Quantization (VQ) embedding as a distinct topic representation, our method leverages both the VQ embedding and its corresponding codebook as an expanded word feature, enabling extraction of a variable number of topics decoupled from the VQ embedding count. 

\paragraph{Topic models with Embedding.}
PCAE~\cite{tu2023fet} also proposes a flexible generation of the output by VAE, which shares a similar idea, and we focus on VQ embeddings as well.
Attempts to include word embeddings, mostly using GloVe~\cite{pennington2014glove}, into generative~\cite{petterson2010word,dieng2020topic,duan2021sawtooth} or non-generative~\cite{wang2022representing,xu2022hyperminer,tu2023fet} topic modeling frameworks have also demonstrated successfully topic modeling performance. 
Moreover, utilizing pre-trained language models (PLMs) such as BERT~\cite{devlin2018bert}, RoBERTa~\cite{liu2019roberta}, and XLNet~\cite{yang2019xlnet} has emerged as a new trend in mining topic models. Many recent studies have enhanced the modeling performance by observing the relation between K-means clusters and topic embeddings~\cite{sia2020tired}. These studies require post-training steps including TF-IDF~\cite{grootendorst2022bertopic} or modifying of PLM embeddings to lie in a spherical embedding space through autoencoding~\cite{meng2022topic} to mitigate the curse-of-dimensionality. Here, our method re-demonstrates the possibility of handling discretized PLM information in a generative manner without post-processing.

\paragraph{Vector Quantized Latent Embedding.} Since \cite{van2017neural} proposes a discretization method for latent embedding incorporated with the variational autoencoding framework, this quantization technique has become an important block for generation, especially for visual generation ~\cite{razavi2019generating}.
Following the study, subsequent studies~\cite{peng2021generating,esser2021taming,yu2021vector,hu2022global} including text to image multi-modal connection~\cite{gu2022vector,tang2022improved,esser2021imagebart} incorporated with autoregressive generation. In this line of studies, we demonstrate that our method can extract topic context from VQ embeddings encapsulating visual information, and generate reasonable samples, simultaneously.

\section{Empirical Analysis}

We analyze the performance of the TVQ-VAE with two applications: document analysis, which is a classical problem in topic modeling, and image generation to show the example of a much more general form of document generation.

\subsection{Document Analysis}

\paragraph{Dataset.} 
We conduct experiments on two datasets: 20 Newsgroups (\textbf{20NG})~\cite{lang1995newsweeder}, the New York Times-annotated corpus (\textbf{NYT})~\cite{sandhaus2008new}, as following the experiments of \cite{dieng2020topic}.
We present the detailed statistics of the datasets in Table~\ref{table:dataset_stats}. While documents in 20NG consist of about $46$ words on average, we note that NYT is a much larger dataset compared to the 20NG dataset, consisting of $32$K documents with $328$ words per document on average.

\paragraph{Baseline Methods.}
To facilitate a comprehensive comparison, we select four representative topic models to encompass BoW-based, embedding-based, neural network-ignored, and neural-network employed approaches as well as generative and non-generative models, as: (1) \textbf{LDA}~\cite{blei2003latent} - a textbook method of BoW-based generative topic model, (2) \textbf{ProdLDA}~\cite{srivastava2017autoencoding} - a BoW-based generative neural topic model (NTM) (3) \textbf{ETM}~\cite{dieng2020topic} - a generative NTM considering Word2Vec embedding~\cite{petterson2010word} as well, (4) \textbf{BerTopic}~\cite{grootendorst2022bertopic} - a non-generative PLM-based topic model utilizing sentence-Bert~\cite{reimers2019sentence} information. We use the implementation from OCTIS~\cite{terragni2021octis} for LDA, ProdLDA, and ETM. For ETM, we use Google's pre-trained Word2Vec as its embedding vector. For BerTopic, we use the official author's implementation.
For \textbf{TVQ-VAE}, we set the embedding number and expansion $k$ to $300$ and $5$.

\paragraph{Implementation Detail. }
To transform words in sentences into vectorized form, we employ Sentence-Bert~\cite{reimers2019sentence}, which converts each word to a $768$-dimensional vector $x$. We use the autoencoder component of VQ-VAE from \cite{meng2022topic}.
The encoder and decoder of the VQ-VAE are composed of a sequence of fully-connected (FC) layers followed by ReLU activation, having intermediate layer dimensions to $[500, 500, 1000, 100]$ and $[100, 1000, 500, 500]$. Consequently, we compress the input vectors to a $100$ dimensional latent vector.

\begin{table}[t]
\small
\centering
\resizebox{0.92\linewidth}{!}
{
\begin{tabular}{lllll}
\toprule
  & \# Doc. & \# Vocab. & \# Total words & \# Labels \\
\midrule
20NG               &16.3K & 1.60K & 0.75M & 20\\
NYT                  &32.0K & 28.7K & 10.5M & 10 (9)\\
\bottomrule
\end{tabular}
}
\caption{Statistics of datasets. For 20NG, we follow the OCTIS setting from \cite{terragni2021octis}. NYT dataset has two categories corresponding to locations (10) and topics (9).}
\label{table:dataset_stats}
\end{table}

\begin{figure*}[t]
    \centering
    \begin{subfigure}[t]{0.32\linewidth}
        \includegraphics[width=0.95\linewidth]{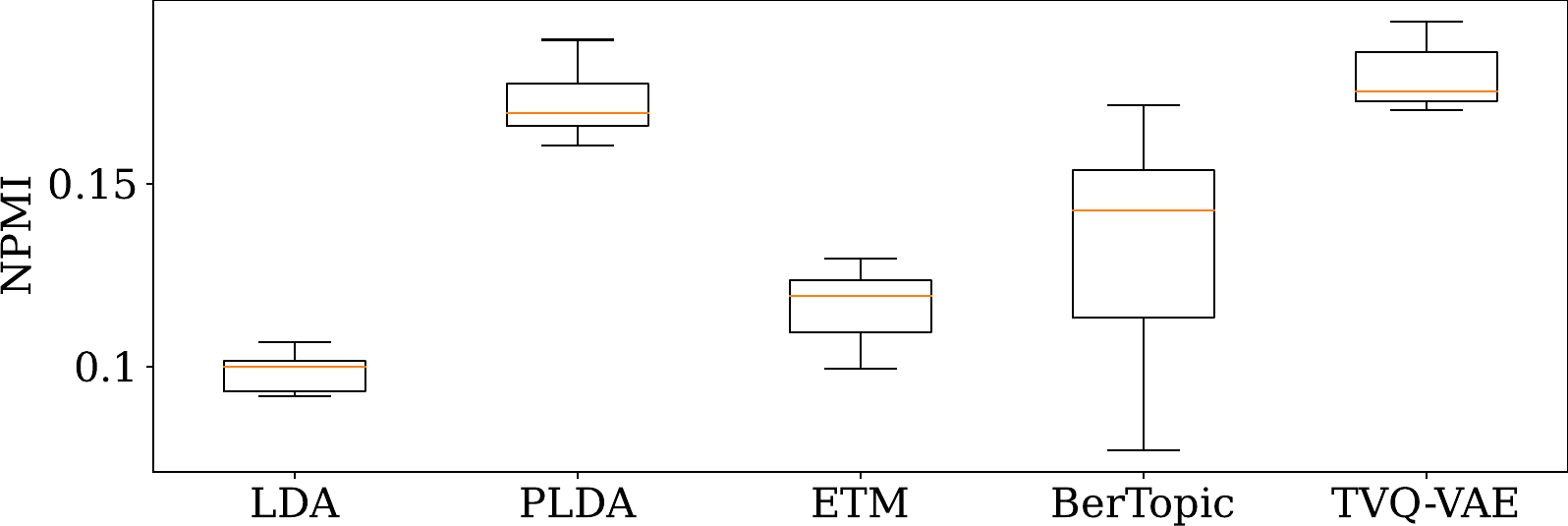}
        \centering
        \caption{20NG-NPMI.}
        \label{fig:20ng_npmi}
    \end{subfigure}
    \begin{subfigure}[t]{0.32\linewidth}
        \includegraphics[width=0.95\linewidth]{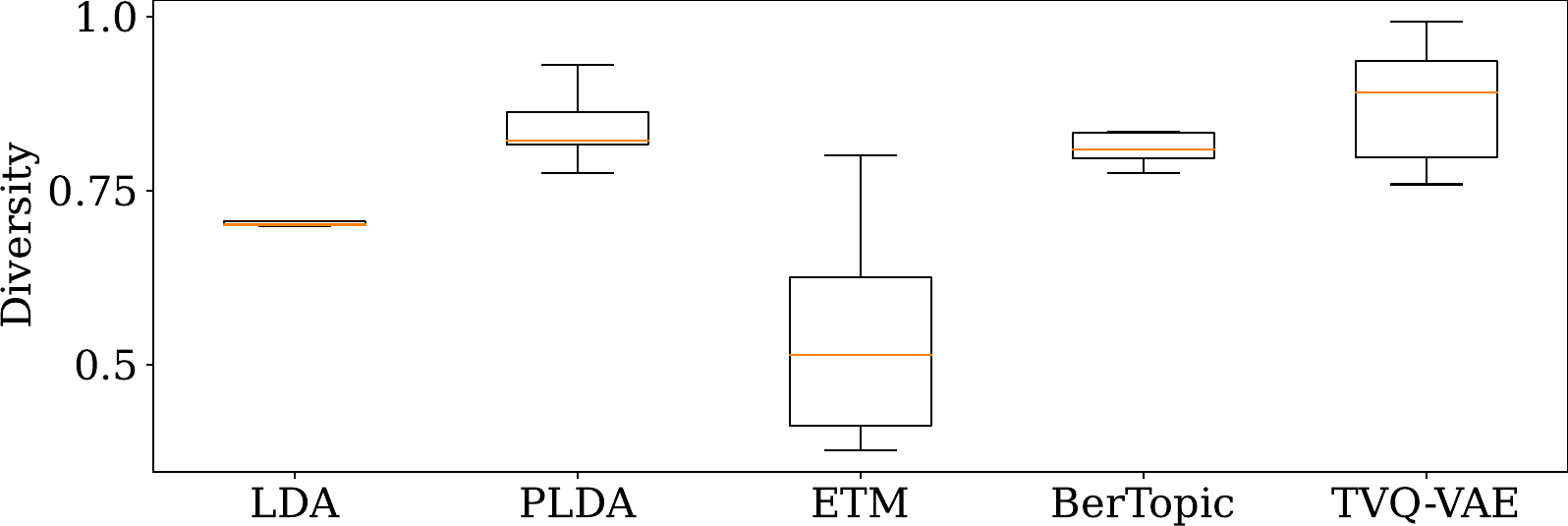}
        \centering
        \caption{20NG-Diversity.}
        \label{fig:20ng_diversity}
    \end{subfigure}
    \begin{subfigure}[t]{0.32\linewidth}
        \includegraphics[width=0.95\linewidth]{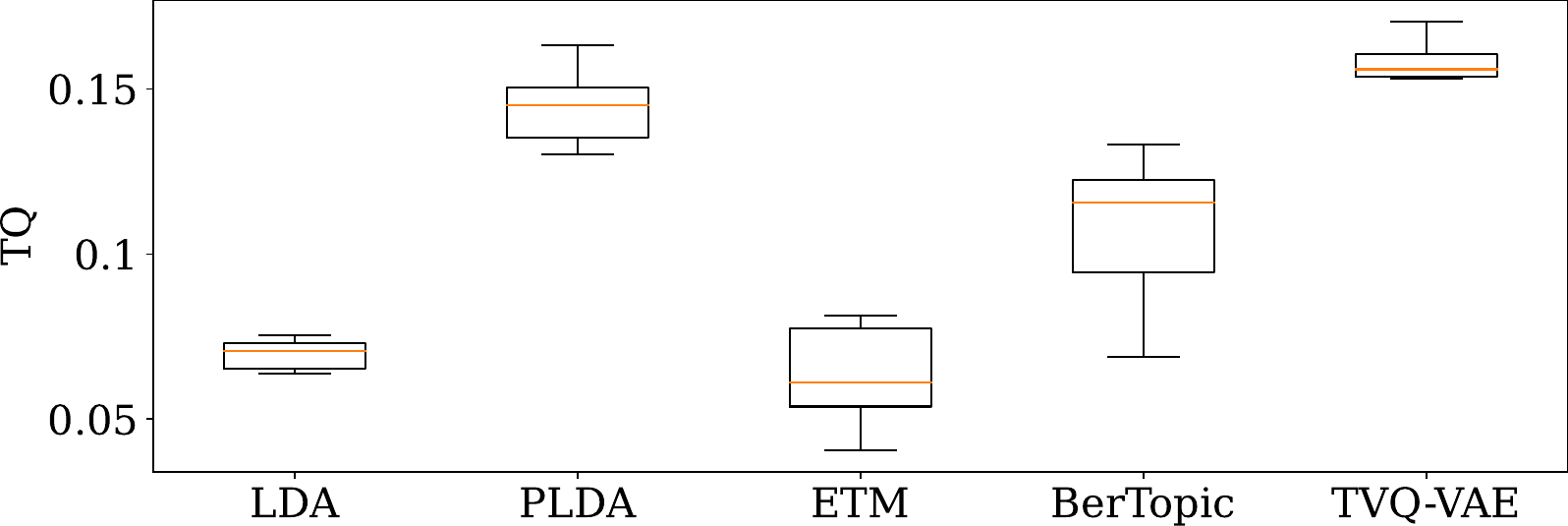}
        \centering
        \caption{20NG-TQ.}
        \label{fig:20ng_tq}
    \end{subfigure}
    \begin{subfigure}[t]{0.32\linewidth}
        \includegraphics[width=0.95\linewidth]{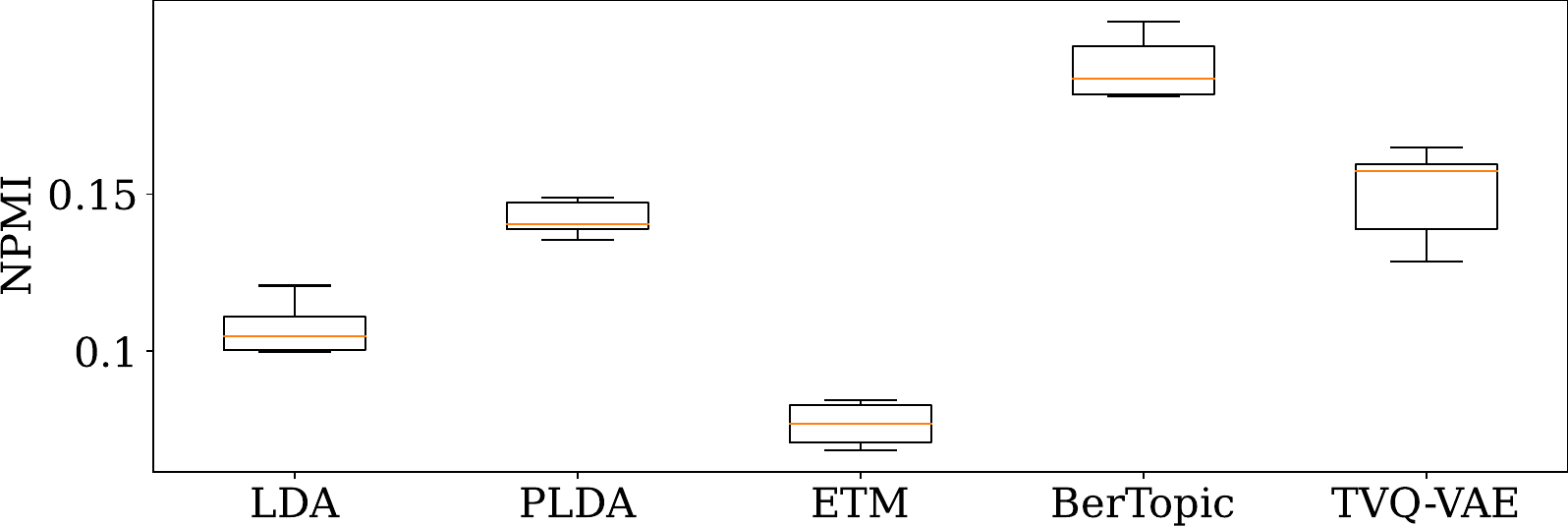}
        \centering
        \caption{NYT-NPMI.}
        \label{fig:nyt_npmi}
    \end{subfigure}
    \begin{subfigure}[t]{0.32\linewidth}
        \includegraphics[width=0.95\linewidth]{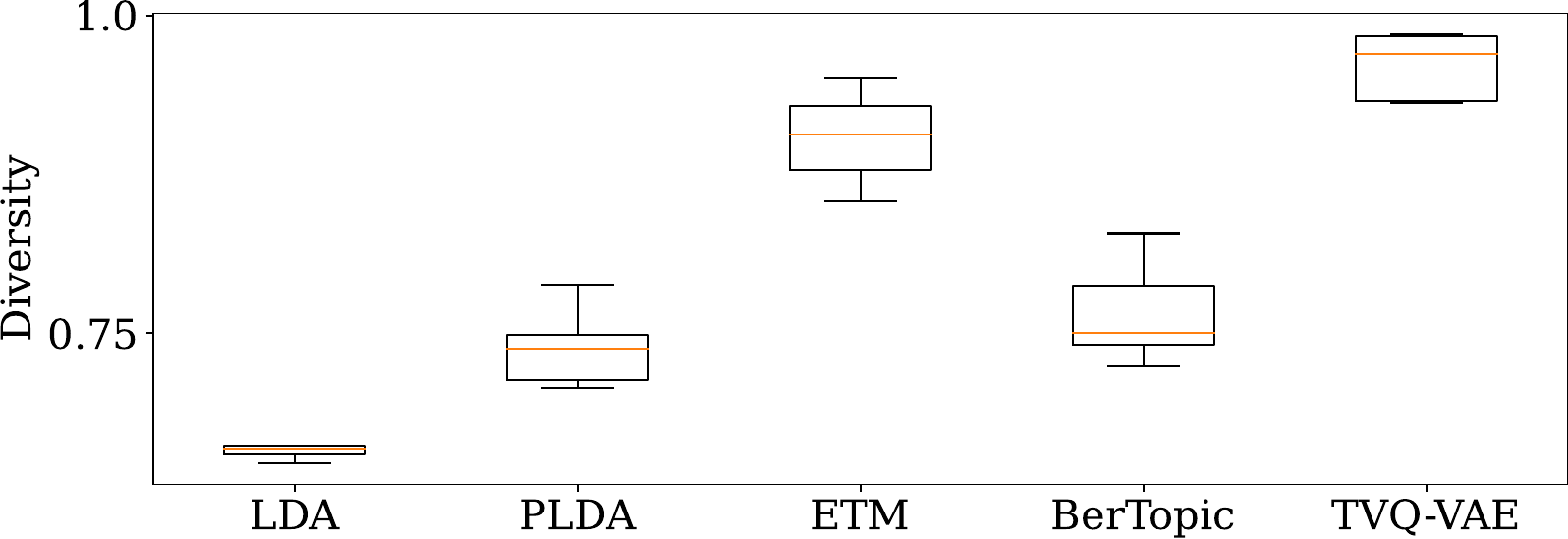}
        \centering
        \caption{NYT-Diversity.}
        \label{fig:nyt_divetsity}
    \end{subfigure}
    \begin{subfigure}[t]{0.32\linewidth}
        \includegraphics[width=0.95\linewidth]{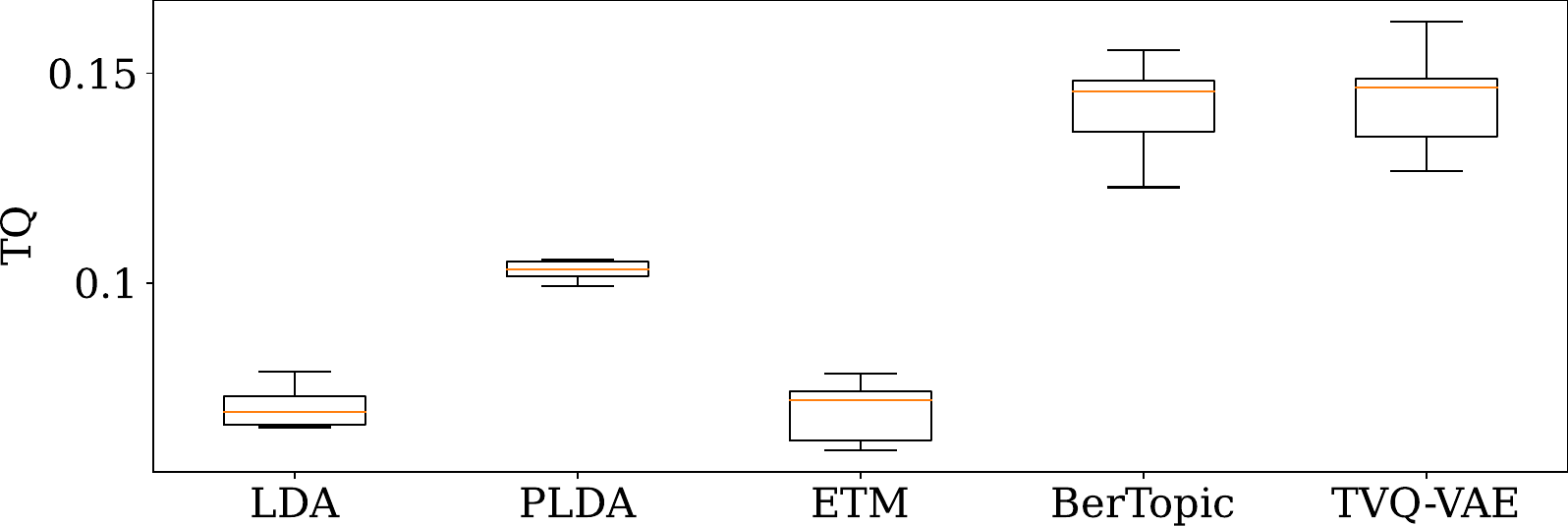}
        \centering
        \caption{NYT-TQ.}
        \label{fig:nyt_tq}
    \end{subfigure}
    \caption{The quantitative evaluation of topic quality over two datasets: 20NG and NYT. The baseline methods are listed from Left to right: LDA, ProdLDA (PLDA), ETM, BerTopic, and TVQ-VAE.}
    \label{fig:tq_eval}
\end{figure*}

\begin{table*}[t]
\small
\centering    
\begin{tabular}{l||cccc|ccc}
\toprule
  & LDA & PLDA & ETM & TVQ-VAE (W) &BerTopic & TopClus & TVQ-VAE\\
\midrule
20NG & (0.309/0.231) & (0.276/0.184) & (\textbf{0.331}/0.207) & (0.222/\textbf{0.249}) &(0.133/0.178) & (0.168/0.221) & (0.210/\textbf{0.242})\\
NYT & (0.144/0.399) & (0.107/0.367) & (0.094/0.346) & (0.176/0.489) & (0.155/0.481) & (0.137/0.461) & (\textbf{0.184}/\textbf{0.510})\\
\bottomrule
\end{tabular}
\caption{Evaluation on Km-NMI and Km-Purity on 20NG and NYT datasets: (Km-NMI / Km-Purity). We note that BerTopic, TopClus and TVQ-VAE both use PLM~\cite{reimers2019sentence}. TVQ-VAE (W) uses Word2Vec instead of the PLM.
    }
\label{table:clustering}
\end{table*}

\begin{figure*}[t]
\begin{center}
    \begin{subfigure}[t]{0.32\linewidth}
        \includegraphics[width=0.95\linewidth]{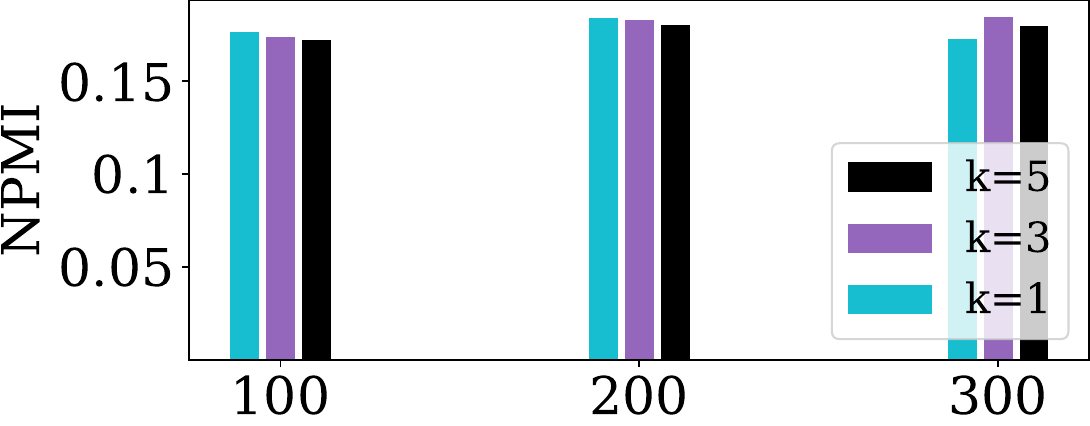}
        \centering
        \caption{20NG-NPMI.}
        \label{fig:abl_20NG_NPMI1}
    \end{subfigure}
    \begin{subfigure}[t]{0.32\linewidth}
        \includegraphics[width=0.95\linewidth]{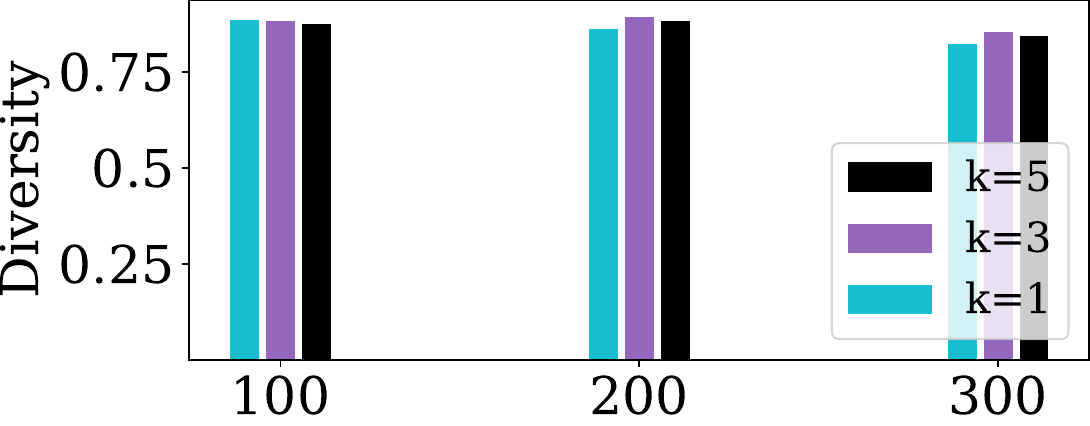}
        \centering
        \caption{20NG-Diversity.}
        \label{fig:abl_20NG_NPMI2}
    \end{subfigure}
    \begin{subfigure}[t]{0.32\linewidth}
        \includegraphics[width=0.95\linewidth]{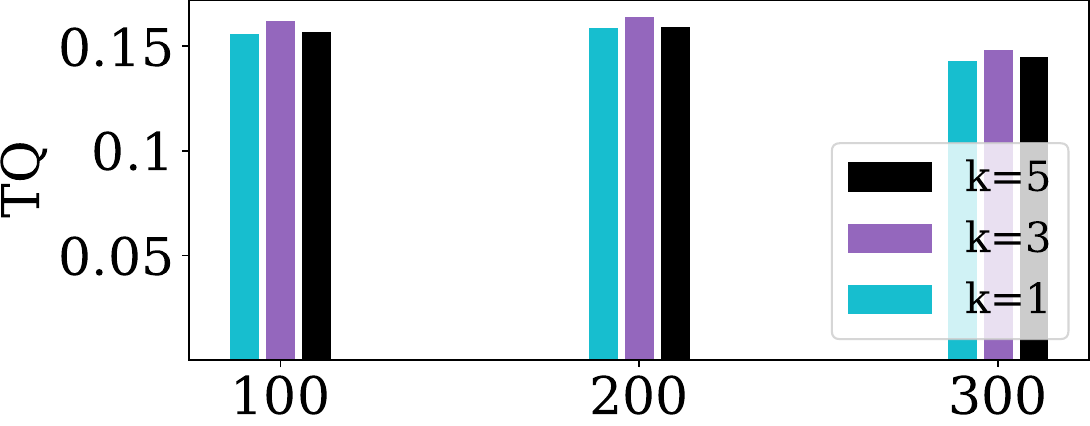}
        \centering
        \caption{20NG-TQ.}
        \label{fig:abl_20NG_NPMI3}
    \end{subfigure} 
    \begin{subfigure}[t]{0.32\linewidth}
        \includegraphics[width=0.95\linewidth]{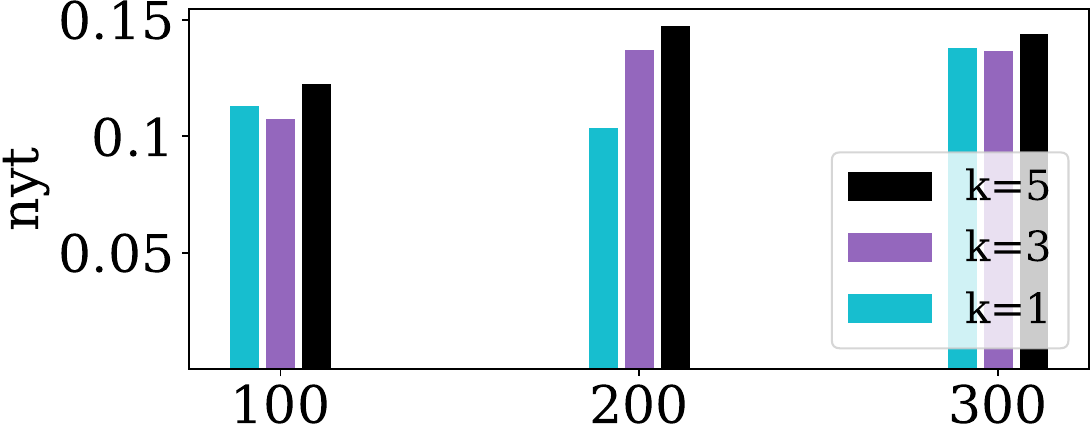}
        \centering
        \caption{NYT-NPMI.}
        \label{fig:abl_20NG_NPMI4}
    \end{subfigure}
    \begin{subfigure}[t]{0.32\linewidth}
        \includegraphics[width=0.95\linewidth]{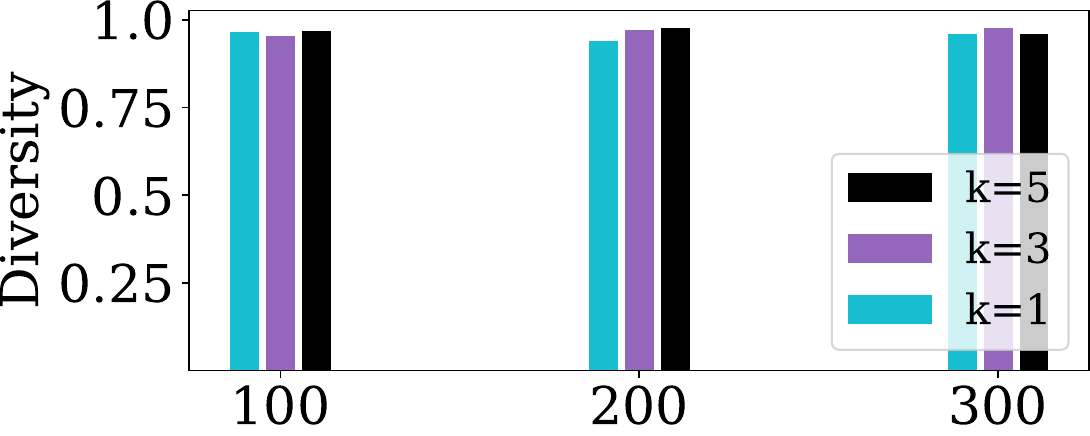}
        \centering
        \caption{NYT-Diversity.}
        \label{fig:abl_20NG_NPMI5}
    \end{subfigure}
    \begin{subfigure}[t]{0.32\linewidth}
        \includegraphics[width=0.95\linewidth]{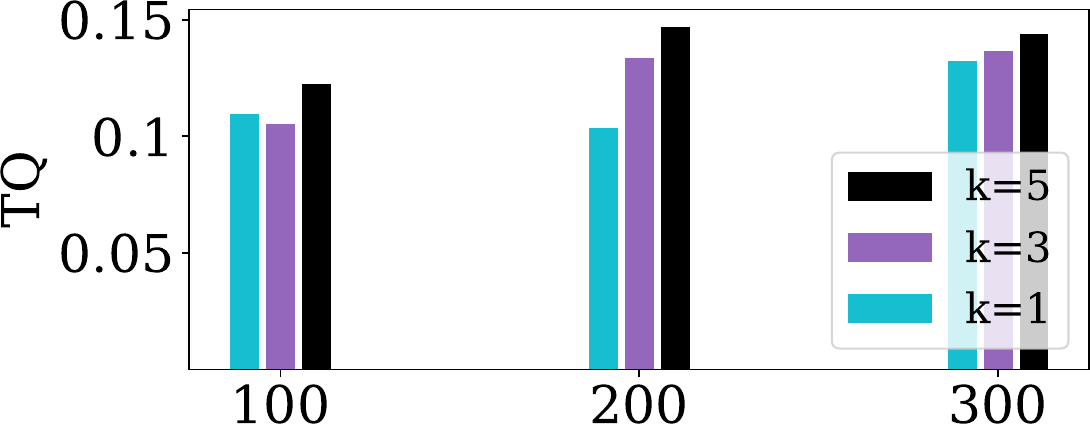}
        \centering
        \caption{NYT-TQ.}
        \label{fig:abl_20NG_NPMI6}
    \end{subfigure} 
\end{center}
\caption{Demonstration of the TQ over various numbers of codebook $\{100, 200, 300\}$ and expansion $k=\{1,3,5\}$.}
\label{fig:tq_ablation}
\end{figure*}

The $NN(\boldsymbol{c})$ of the Algorithm 2, which draws $\boldsymbol\theta$, of the main manuscript are implemented using the inference network architecture of ProdLDA~\cite{srivastava2017autoencoding}, as implemented in OCTIS~\cite{terragni2021octis}.
The $NN(\boldsymbol{c})$ is implemented by three consecutive linear layers followed by tangent hyperbolic activation, which has latent dimensions to $[100,100]$.
We pretrained the VQ-VAE architectures for 20 epochs and trained the remaining parts of TVQ-VAE for 200 epochs with by optimizer~\cite{kingma2014adam} with a learning rate of $5 \times 10^{-3}$. The batch size was set to $256$ for both training and pretraining.

\paragraph{Evaluation Metric.}
We evaluate the model's performance in terms of topic quality (TQ) and document representation, following the established evaluation setup for topic models. TQ is evaluated based on Topic Coherence~(TC) and Topic Diversity~(TD). TC is estimated by using Normalized Point-wise Mutual Information (NPMI)~\cite{aletras2013evaluating}, quantifying the semantic coherence of the main words within each topic. NPMI scores range from $-1$ to $1$, with higher values indicating better interpretability. TD measures word diversity by computing the unique word numbers among the top $25$ words across all topics~\cite{dieng2020topic}. TD scores range from $0$ to $1$, with higher values indicating richer word diversity. TQ is defined as the multiplication of the TC, measured by NPMI, and TD values.

Furthermore, to measure document representation, we report the purity and Normalized Mutual Information (NMI)~\cite{schutze2008introduction}. Following \cite{xu2022hyperminer}, we cluster the $\theta_d$ of every document $d$ and measure the purity and NMI termed as \textbf{Km-NMI} and \textbf{Km-Purity}. Both values range from $0$ to $1$, and the higher values indicate better performance.

\paragraph{Topic Quality Evaluation.} We present the evaluation results for topic quality (TQ), as depicted in Figure~\ref{fig:tq_eval}. From the evaluation settings outlined in \cite{grootendorst2022bertopic}, we infer a range of 10 to 50 topics with a step size of 10 and measure their TC and TD to evaluate TQ. 

First, we evaluate the performance of TVQ-VAE on the 20NG dataset, which is widely used in the field of topic modeling.
Notably, the TVQ-VAE demonstrates either comparable or superior performance compared to other baselines in terms of TQ measures. It is worth mentioning that the 20NG dataset has a  small vocabulary size, which stands at $1.6K$. This scale is considerably smaller considering the number of TVQ-VAE codebook sizes. 
These results represent that TVQ-VAE effectively extracts topic information for documents with limited size, where BoW-based topic models like ProdLDA have exhibited impressive success.

In the NYT dataset, characterized by a significantly larger vocabulary to 20NG, the TVQ-VAE model achieves competitive topic quality when utilizing only $300$ virtual codebooks, which accounts for less than $1\%$ of the original vocabulary size. Among the baselines, BerTopic stands out as it demonstrates exceptional performance, particularly in terms of NPMI, deviating from the results observed in the 20NG dataset. The result verifies BerTopic's claim that PLM-based methods are scalable for larger vocabulary. 

\begin{table*}[t]
\small
\centering    
\resizebox{0.92\linewidth}{!}
{
\begin{tabular}{ll}
\toprule
 20NG & NYT \\
\midrule
muslim, turkish, armenian, arab, country &  gubernatorial, campaign, democratic, race,  election\\
archive, server, entry, mail, resource & japan, Japanese, tokyo, hokkaido, fuji\\
graphic, internet, database, programming, guide & spacecraft, orbit, module, capsule, endeavour\\
president, decision, meeting, yesterday, administration & administration, nato, pluto, washington, nuclear\\
gun, violent, accident, crime, risk & military, american, pentagon, command, force\\
voltage, circuit, output, wire, frequency & school, kindergarten, mathematics, education, elementary\\
graphic, internet, database, programming, guide & bank, investment, firm, supervisory, stock\\
player, average, career, league, pitcher & european, monetary, germany, west, union\\
shipping, sale, sell, brand, offer & senate, legislation, republican, procedural, congress\\
existence, belief, argument, atheist, conclusion & waterfront, marina, park, center, downtown\\
gay, sex, homosexual, moral, sexual & growth, percent, quarter, year, economy\\

\bottomrule
\end{tabular}
}
\caption{Topic Visualization of TVQ-VAE. We demonstrate top 5 words for each topic.
    }
\label{table:topic_vis}
\end{table*}

\begin{table}[t]
\small
\centering    
\begin{tabular}{c|l}
\toprule
Codebook 106 & money, profit, guarantee, hope, financial, ...\\
\midrule
Codebook 176 & life, today, time, economy, bank, ...\\
\midrule
Codebook 207 & two, zero, sixth, asset, analyst, ... \\
\bottomrule
\end{tabular}
\caption{Conceptual-word to word mapping in NYT dataset.}
\label{table:conceptual_word}
\end{table}

Figure~\ref{fig:tq_ablation} presents the ablation study conducted with varying the number of codebooks by $\{100,200, 300\}$ and the expansion values by $k=\{1,3,5\}$.
In the case of the 20NG dataset, the evaluation results indicate minimal performance differences across all settings. This presents that the choice of embedding and expansion numbers does not necessarily guarantee performance enhancements. This may happen due to the relatively small vocabulary size of 20NG, Moreover, exceeding certain bounds for the number of codebooks and expansion appears to capture no additional information from the original dataset.
Conversely, the evaluation results obtained from the NYT dataset support our analysis. Here, the performance improves with larger codebook sizes and expansion numbers, given the vocabulary size of approximately 20 times that of the 20NG.

\paragraph{Document Representation Evaluation.}
Table~\ref{table:clustering} presents the km-NMI and km-Purity scores for each topic model. In the 20NG dataset, characterized by a relatively smaller vocabulary size, the previous BoW-based method exhibited superior NMI scores. However, in the case of the NYT dataset, PLM-based methods like BerTopic and TVQ-VAE demonstrated higher performance. We additionally evaluate TopClus~\cite{meng2022topic} as a variant of the PLM-based topic model. These findings suggest that our TVQ-VAE model exhibits robust document representation capabilities, particularly as the vocabulary size expands. 

Additionally, when employing Word2Vec with TVQ-VAE, we observed performance on par with that of PLM-based TVQ-VAE. In fact, in the case of the 20NG dataset, Word2Vec-based TVQ-VAE even exhibited superior performance. We hypothesize that this outcome can be attributed to the comparatively reduced number of words and vocabulary in the 20NG dataset when compared to NYT. This observation aligns with a similar trend noticed in ETM, which also utilizes Word2Vec.

We also note that PLMs like BerTopic excel on larger datasets such as NYT, but not on smaller ones like 20NG, suggesting that PLMs' breadth may not translate to depth in constrained datasets. In the smaller datasets, the model's broad lexical coverage may result in singular categories with high purity but restricted breadth, thereby diminishing Km-NMI. TopClus results corroborate this, underscoring the influence of the data set size on the model efficacy.

\paragraph{Topic and Codebook Demonstration.} 
Table~\ref{table:topic_vis} provides a visual summary of the top 5 representative words associated with each topic in both the 20NG and NYT datasets. It is evident from this table that the words within each topic exhibit clustering behavior, indicating a shared semantic similarity among them.
Also, we show that the conceptual codebook functions as a semantic cluster, aggregating words with higher semantic proximity just before topic-level clustering. The example showcasing the collection of words for each codebook illustrates this tendency, in Table~\ref{table:conceptual_word}.

\subsection{Image Generation}

\begin{figure*}[t]
\begin{center}
    \begin{subfigure}[t]{0.48\linewidth}
        \includegraphics[width=0.95\linewidth]{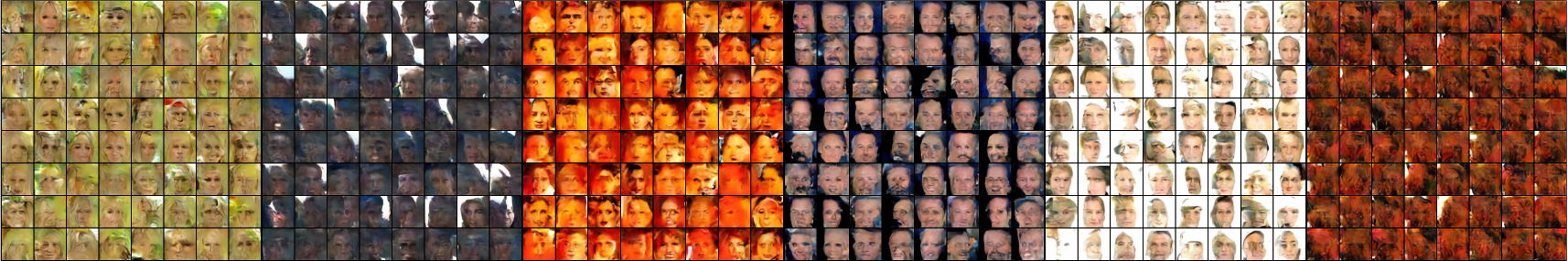}
        \centering
        \caption{Topic visualizations on CelebA dataset.}
        \label{fig:fig_topics_celeba}
    \end{subfigure}
    \begin{subfigure}[t]{0.48\linewidth}
        \includegraphics[width=0.95\linewidth]{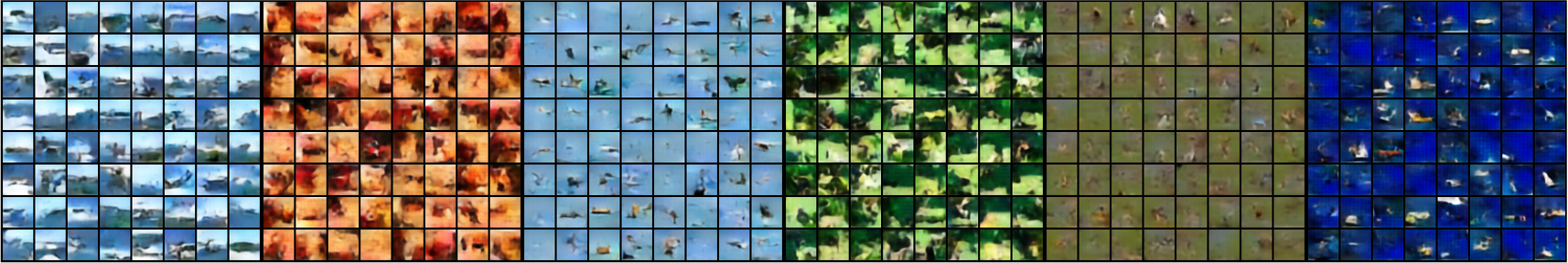}
        \centering
        \caption{Topic visualizations on CIFAR-10 dataset}
        \label{fig:fig_topics_cifar}
    \end{subfigure}
    \begin{subfigure}[t]{0.48\linewidth}
        \includegraphics[width=0.95\linewidth]{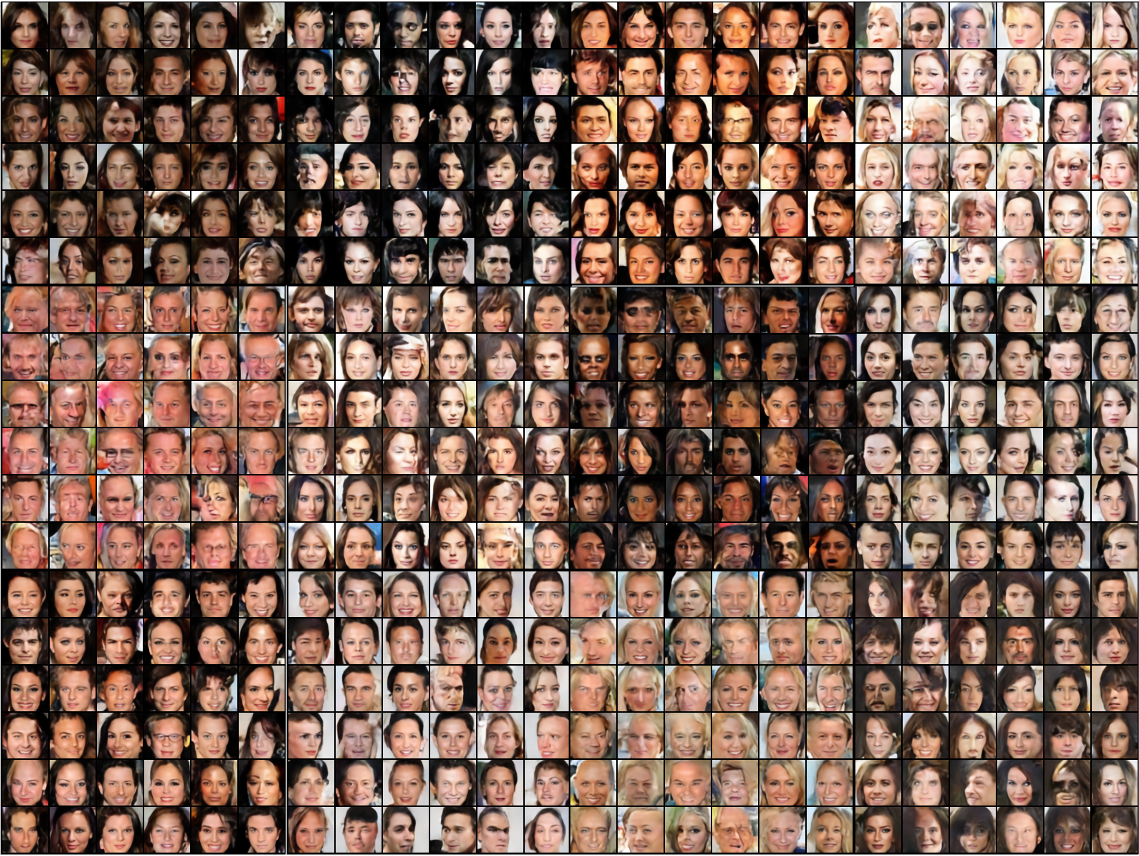}
        \centering
        \caption{Reference-based generation on CelebA dataset.}
        \label{fig:fig_i2i_celeb}
    \end{subfigure}
    \begin{subfigure}[t]{0.48\linewidth}
        \includegraphics[width=0.95\linewidth]{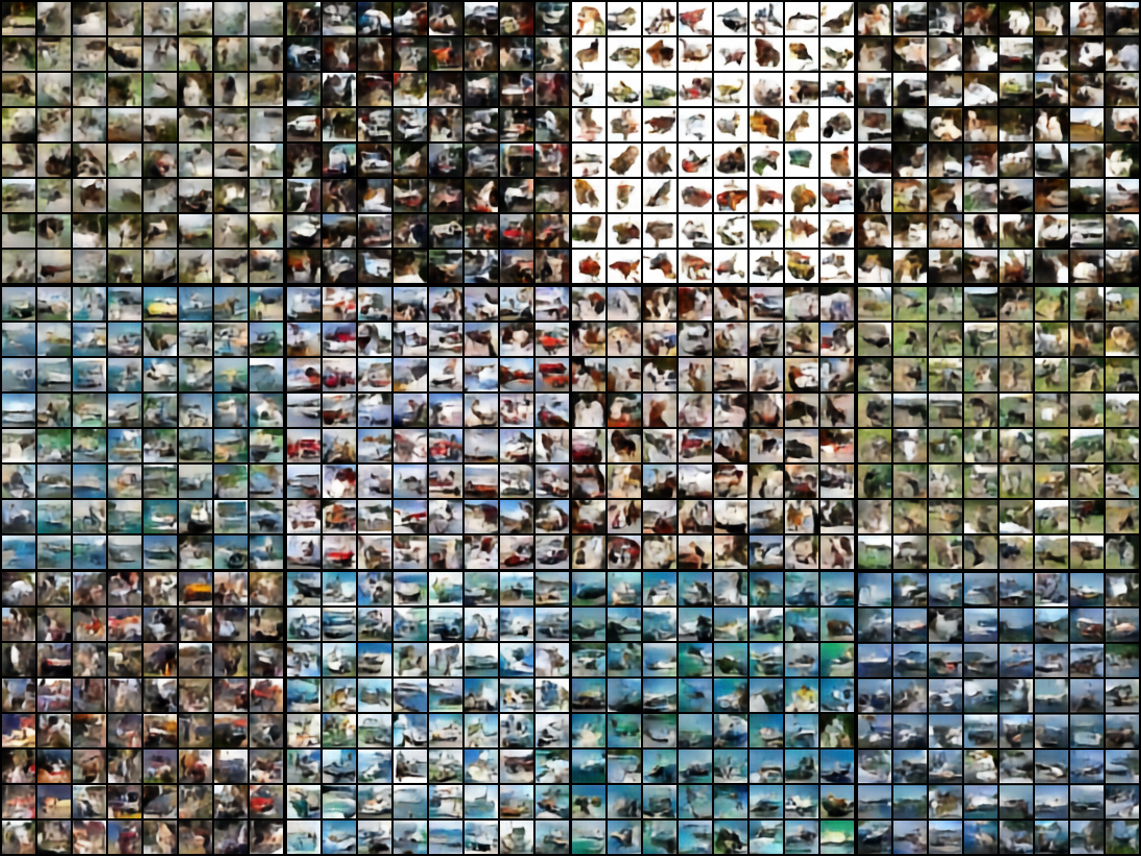}
        \centering
        \caption{Reference-based generation on CIFAR-10 dataset}
        \label{fig:fig_i2i_cifar10}
    \end{subfigure}

\end{center}
\caption{Illustrations of visualized topics and reference-based generation for topic number $K$ of $100$, from TVQ-VAE (P).}
\label{fig:topic_generation}
\end{figure*}

\paragraph{Dataset.} To demonstrate that TVQ-VAE can mine topic information from the visual codebooks from VQ-VAE, we first tested our method into two image datasets: CIFAR-10~\cite{krizhevsky2009learning} and CelebA~\cite{liu2015faceattributes} typically used for supervised and unsupervised image generation, respectively. While CIFAR-10 contains $60K$ $32$x$32$ dimensional images with 10 class objects, CelebA consists of about $200K$ of annotated facial images. We center-crop and resize the images to have $64$x$64$ dimension. We convert the images to a sequence consisting of $64$ and $256$ codebooks, respectively, i.e., each image is represented as a document having $64$ and $256$ words. 
Also, to validate the proposed method TVQ-VAE into larger resolution image, we used FacesHQ~\cite{esser2021taming} dataset, which includes FFHQ~\cite{karras2019style} and CelebaHQ~\cite{karras2017progressive} datasets.

\paragraph{Baseline Methods.} Since the general form of document generation conditioned to a topic is a newly proposed task, it is difficult to directly compare to the previous methods. Quantitatively, therefore, we compare the TVQ-VAE to the baseline VQ-VAE generation guided by PixelCNN prior, TVQ-VAE (P), which is a typical method of auto-regressive generation. All the network architecture of the VQ-VAE and PixelCNN is equivalent to those in TVQ-VAE. Also, we apply the TVQ-VAE concept into \cite{esser2021taming}, which is a representative AR method using Transformer and VQ-codebooks, abbreviated as TVQ-VAE (T) and test with FacesHQ dataset.

\paragraph{Evaluation.} Regarding the quantitative evaluation, we utilize the Negative Log-Likelihood (NLL) metric on the test set, a widely adopted measure in the field of auto-regressive image generation. A lower NLL value means better coverage of the dataset.
For the qualitative evaluation, we demonstrate the generated images corresponding to each topic, illustrating the topic's ability to serve as a semantic basis in shaping the generated data. Furthermore, we show image generation examples conditioned on a reference image by leveraging its topic information expressed as $\theta$.

\paragraph{Implementation Detail. }
We employed the TVQ-VAE (P) framework, utilizing VQ-VAE and PixelCNN architectures from a well-known PyTorch repository\footnote{https://github.com/ritheshkumar95/pytorch-vqvae.git}. The VQ-VAE model integrates 64 and 256 codebooks for 32x32 and 64x64 image resolutions, respectively. 
Its encoder features four convolutional (Conv) blocks: two combining Conv, batch normalization (BN), and ReLU activation, and two residual blocks with Conv structures outputting dimensions of $256$. 
The latent vector dimensions are likewise set to $256$. The decoder comprises two residual and two ConvTranspose layers with intermediate channels to $256$, using ReLU activations.

For topic information extraction, we use an inference network $NN(\boldsymbol{c})$, equivalent to that in Document analysis. We conditional embedding of the GatedCNN architecture to get topic embedding $(\theta_d\cdot\hat{\boldsymbol\beta}\cdot\hat{\boldsymbol\rho})$ instead of the original class-conditional embedding. 
For pretraining the VQ-VAE, we employ the Adam optimizer for $100$ epochs with a learning rate of $2 \times 10^{-4}$. Similarly, in TVQ-VAE(P), the topic modeling and PixelCNN prior are trained for $100$ epochs using an identical optimizer setup and a batch size of $128$.

\begin{figure*}[t]
\begin{center}
    \includegraphics[width=0.95\linewidth]{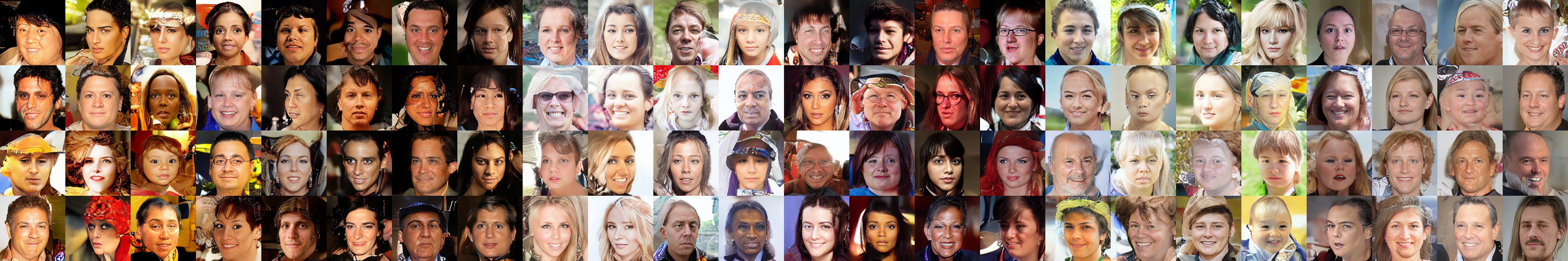} 
\end{center}    
\caption{Illustrations of reference-based generation applying TVQ-VAE (T) for topic number $K$ of $100$.}
\label{fig:topic_generation_transformer}
\end{figure*}

\begin{figure}[t]
\begin{center}
\includegraphics[width=0.95\linewidth]{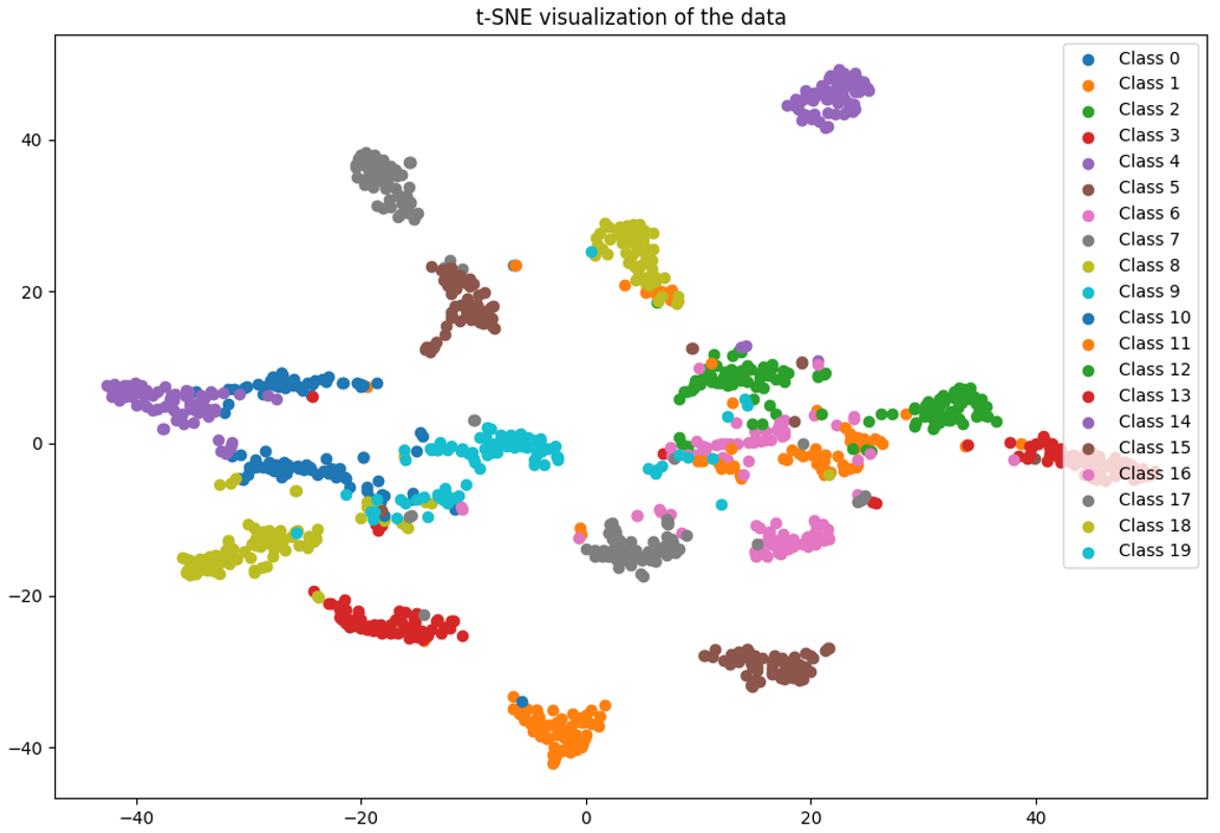}
\end{center}
\caption{Visualization of topic embedding by t-SNE, from TVQ-VAE (P) for CIFAR-10 generation, 512 codebooks.}
\label{fig:visualization_tsne_codebook}
\end{figure}

Furthermore, the proposed TVQ-VAE was extended to TVQ-VAE (T) by applying a representative AR model from \cite{esser2021taming}, using Transformer and VQ-codebooks from VQGAN, to generate high-resolution images as the topic-driven documents. 
TVQ-VAE (T) facilitates codebook generation for context-rich visual parts through convolutional layers and enables auto-regressive prediction of codebook indices using Transformer. Topic information extraction is performed through an inference network in the same manner as previously described.

To reflect topic information to the Transformer, each codebook token was augmented with the topic embedding $(\theta_d\cdot\hat{\boldsymbol\beta}\cdot\hat{\boldsymbol\rho})$ to integrate topic information. This augmented embedding becomes an additional input for Transformers, minGPT architecture from Karpathy\footnote{https://github.com/karpathy/minGPT}.
We use the pre-trained VQGAN codebook for the FacesHQ dataset from the official repository of \cite{esser2021imagebart}. 

Specifically, we use the topic embedding for two purposes, one for augmented token and the other for bias for the input of the transformer block, consisting of the causal self-attention layer. As the augmented token, we repeatedly assign a $256$ number of topic tokens before the image tokens, where the number is $256$, also. Furthermore, for each transformer block output that has a $512$ token length, we add the topic tokens as a bias for the latter $256$ tokens, which is the predicted image token of the block.
We repeatedly expand the topic embedding dimension to $1024$ from the original $256$, to align the dimension size to those of the image token.

\paragraph{Quantitative Evaluation.}
Table~\ref{table:quant_image} presents the NLL evaluation results for the CelebA and CIFAR-10 datasets. 
We conjecture that the extraction of the topic variables $\theta$ and $\beta$ helps the easier generation of the samples, quantified by lower NLL, since the topic variables already extract the hidden structures of the dataset which is originally the role of the generation module.
The evaluations conducted on the CelebA and CIFAR-10 datasets yield contrasting outcomes. Specifically, in the case of CelebA, the unsupervised baseline exhibits a lower NLL. Conversely, for CIFAR-10, the NLL demonstrates a linear decrease with an increasing number of topics, surpassing the NLL values of both unsupervised and class-label supervised generation methods.

The complexity of the two datasets provides insights into the observed patterns. The CelebA dataset comprises aligned facial images, and the preprocessing step involves center-cropping the facial region to produce cropped images that specifically include the eyes, nose, and mouth. This preprocessing step effectively reduces the dataset's complexity. In contrast, the CIFAR-10 dataset consists of unaligned images spanning ten distinct categories, resulting in an increased level of complexity. Previous evaluations from the baseline methods~\cite{van2016pixel,van2017neural} have similarly highlighted the challenging nature of NLL-based generation for CIFAR-10.
Therefore, we contend that the evaluation in Table~\ref{table:quant_image} supports our conjecture that topic extraction can enhance the model's generation capabilities for complicated datasets. especially for complicated datasets.

\begin{table}[t]
\small
\centering    
\begin{tabular}{lcccccc}
\toprule
  & U & 10 & 20 & 50 & 100 & S\\
\midrule
CelebA &3.10 & 3.15 & 3.14 & 3.14 & 3.13 & -\\
CIFAR-10 & 3.29 & 3.27 & 3.25 & 3.22 & 3.20 & 3.29\\
\bottomrule
\end{tabular}
\caption{NLL evaluation on CelebA and CIFAR-10 dataset. The terms `U' and `S' denote unsupervised and supervised generation from the VQ-VAE integrated with PixelCNN prior. The numbers $\{10, 20, 50, 100\}$ denote the number of topics assigned to TVQ-VAE.
    }
\label{table:quant_image}
\end{table}

\paragraph{Qualitative Evaluation.}
Figure~\ref{fig:topic_generation} shows visual examples of topics as well as generated samples obtained from reference images from TVQ-VAE (P). The visualized topic examples in Figures~\ref{fig:fig_topics_celeba} and \ref{fig:fig_topics_cifar}, arranged in an $8\times 8$ grid, illustrate the generated samples obtained by fixing $\theta$ in Equation~(\ref{eq:tvq_gen_ar}) to a one-hot vector corresponding to the topic indices. Subsequently, the PixelCNN prior $p_{pix}(\cdot|\boldsymbol\theta\cdot\hat{\boldsymbol\beta}\cdot\hat{\boldsymbol\rho})$ generates the codebook sequences by an auto-regressive scheme, conditioned on each $k$-th topic vector $\rho_{(\beta)} = \beta_k\cdot\hat{\boldsymbol\rho}$.
The topic visualization shows that each topic exhibits distinct features, such as color, shape, and contrast.

Furthermore, we demonstrate the generation ability of the TVQ-VAE (P) by first, extracting the topic distribution $\theta_d$ of the image $x_d$, and subsequently generating new images from the extracted $\theta_d$. In this case, we expect the newly generated images to share similar semantics to the original image $x$, which is called \textbf{reference-based generation}.
As shown in Figures~\ref{fig:fig_i2i_celeb} and \ref{fig:fig_i2i_cifar10}, we generate images similar to the reference image, which is on the top-left corners each. The visual illustration for both CIFAR-10 and CelebA clearly demonstrates that TVQ-VAE (P) effectively captures the distinctive attributes of reference images and generates semantically similar samples by leveraging the integrated topical basis.

Figure~\ref{fig:topic_generation_transformer} demonstrates the sample generation examples with higher resolution, 256, from the TVQ-VAE (T) trained from FacesHQ dataset, with the equivalent format to the reference-based generation in Figure \ref{fig:topic_generation}. Both cases show that the topic embedding from each reference image captures essential features of the image for generating semantically close images, and the proposed TVQ-VAE method can be effectively applied to two different AR models: PixelCNN (P) and Transformer (T).

\paragraph{Visualization of Embedding Space. }
For more demonstration of the proposed concepts, we present t-SNE~\cite{van2008visualizing} plot for topic embedding space, in Figure~\ref{fig:visualization_tsne_codebook}. Each data point on the plot corresponds to the topic embedding of generated images derived from identical reference sources. This serves as a visual representation of the capability of our TVQ-VAE to produce images that exhibit semantic proximity to their respective reference images. Furthermore, it is evident that the generated images form distinct clusters within the embedding space.

\section{Conclusion and Future Remark}
We introduced TVQ-VAE, a novel generative topic model that utilizes discretized embeddings and codebooks from VQ-VAE, incorporating pre-trained information like PLM. Through comprehensive empirical analysis, we demonstrated the efficacy of TVQ-VAE in extracting topical information from a limited number of embeddings, enabling diverse probabilistic document generation from Bag-of-Words (BoW) style to autoregressively generated images.
Experimental findings indicate that TVQ-VAE achieves comparable performance to state-of-the-art topic models while showcasing the potential for a more generalized topic-guided generation. Future research can explore the extension of this approach to recent developments in multi-modal generation.

\section{Acknowledgements}
We thank Jiyoon Lee\footnote{Independent researcher (jiyoon.lee52@gmail.com). The co-research was conducted during her internship at ImageVision, NAVER Cloud, in 2023.} for the helpful discussion, experiments, and developments for the final published version. This research was supported by the Chung-Ang University Research Grants in 2023 and the Institute of Information \& communications Technology Planning \& Evaluation (IITP) grant funded by the Korean government(MSIT) (2021-0-01341, Artificial Intelligence Graduate School Program (Chung-Ang Univ.)).

\bibliography{aaai24}

\end{document}